\documentclass[12pt]{article}	

\usepackage{amsmath}
\usepackage{amsthm}
\usepackage{amsfonts}
\usepackage{amssymb}
\usepackage{latexsym}
\usepackage{graphicx}
\usepackage{psfrag}
\usepackage{a4}
\usepackage{caption}
\usepackage{float}

\input comment.sty

\usepackage{hyperref}

\numberwithin{equation}{section}

\makeatletter  % Allow the use of @ in command names
\renewenvironment{figure}[1][\fps@figure]{%
	\@float{figure}[#1]%
	\small
}{%
	\end@float
}
\makeatother   % Cancel the effect of \makeatletter

\newcommand{\captionfonts}{\small}
\makeatletter  % Allow the use of @ in command names
\long\def\@makecaption#1#2{%
	\vskip\abovecaptionskip
	\sbox\@tempboxa{{\captionfonts #1 #2 }}       %  Create tempbox of one line: "Fig ... <text>"
	\ifdim \wd\@tempboxa >0.8\textwidth           %  if its length >0.8\textwidth,
	{\begin{center}\parbox[t]{0.8\textwidth}{\captionfonts #1 #2}\end{center}}  % .. put it to the parbox of this width
	\else
	{\begin{center}\captionfonts #1 #2\end{center}\par}                        % .. otherwise print centered.
	\fi
}
\renewcommand{\fnum@figure}{\textbf{Fig.~\thefigure.}}  %  Make bold{Fig No}
\makeatother   % Cancel the effect of \makeatletter

\newcommand{\x}{\mathbf{x}}	
\newcommand{\y}{\mathbf{y}}	
\newcommand{\X}{\mathbf{X}}	
\newcommand{\Y}{\mathbf{Y}}	
\newfont{\mathsets}{msbm10 scaled 1200}
\renewcommand{\Re}{\mbox{\mathsets R}}			%  Real numbers
\newcommand{\ts}{\textsuperscript}

\usepackage[absolute]{textpos}
\begin{document}
\pdfbookmark[1]{Title}{Title}

\begin{textblock}{18}(1.5, 1)
	\noindent\it\centering \mbox{To appear in Proc.\ of 3rd IFAC Workshop on Automatic Control in Offshore Oil and Gas Production}, Esbjerg, Denmark, May 30--June 01, 2018.
\end{textblock}

\title{A Machine Learning Approach for Virtual Flow Metering and Forecasting}
	\author{Nikolai Andrianov\footnote{The Danish Hydrocarbon Research and Technology Centre, 
			Technical University of Denmark, 
			2800 Kgs.\ Lyngby, 	Denmark. E-mail: \href{mailto:nandria@dtu.dk}{nandria@dtu.dk}.}
	}

\date{January 8, 2018}
\maketitle

\begin{abstract}                % Abstract of not more than 250 words.
We are concerned with robust and accurate forecasting of multiphase flow rates in wells and pipelines during oil and gas production. In practice, the possibility to physically measure the rates is often limited; besides, it is desirable to estimate  future values of multiphase rates based on the previous behavior of the system. In this work, we demonstrate that a Long Short-Term Memory (LSTM) recurrent artificial network is able not only to accurately estimate the multiphase rates at current time (i.e., act as a virtual flow meter), but also to forecast the rates for a sequence of future time instants. For a synthetic severe slugging case, LSTM forecasts compare favorably with the results of hydrodynamical modeling. LSTM results for a realistic noizy dataset of a variable rate well test show that the model can also successfully forecast  multiphase rates for a system with changing flow patterns. 
\end{abstract}

%===============================================================================

\section{Introduction}
Accurate multiphase flow rate measurement is an indispensable tool for production optimization from oil and gas fields, especially in an offshore environment (see e.g.~\cite{Amin:2005}).  Currently, there are two industry-accepted solutions for providing such measurements: using test separators and using multiphase flow meters. While these  approaches have their advantages and disadvantages of (see e.g.~\cite{Handbook:2005}), both of them require hardware installations. This can limit the applicability of physical metering devices due to possible transportation issues, space and security considerations, and high costs. 

A virtual flow meter (VFM) is a mathematical model which allows to estimate multiphase rates using available data on the flow. A VFM, primarily using readily available cheap measurements (such as pressure and temperature), can potentially serve as a cost-efficient addition to physical flow metering devices.

VFM models can be classified as hydrodynamical or data-driven. In the hydrodynamical approach one  typically solves  the phase conservation equations in a pipe geometry, which requires the choice of an adequate mathematical model,  appropriate numerical method, and availability of a large number of input data. An advantage of this method is that one can estimate various parameters at arbitrary points of the flowline. A comparison of several hydrodynamical VFMs is presented in~\cite{Amin:2015}. 

The data-driven approach is a system identification tool, which requires the user to accept one of generic model structures. % (e.g.\ statistical analysis or neural networks). 
Such models exploit no prior knowledge on the flow and produce essentially data descriptions. In practice, it is easier to setup a data-driven model as compared to a hydrodynamical one. However, data-driven predictions do not have a physical interpretation and it is not possible to estimate parameters with no historical data. Despite these shortcomings, the use of data-driven VFMs is gaining momentum in the industry, see~\cite{Briers:2016}.

One important difference between hydrodynamical and data-driven VFMs is the ability of the latter not only to \emph{predict} rates (i.e., estimate rates at the current time instant $t_n$), but also to \emph{forecast}  rates at  future time instants $t_{n+1},t_{n+2},\dots$. Indeed, without \emph{a priori} knowledge of time-varying boundary conditions, a hydrodynamical model is only able to yield forecasts at the next time instant $t_{n+1}$.

The goal of the present paper is to evaluate the forecasting capability of a class of data-driven VFMs which use artifical neural networks (ANNs). Feedforward  ANNs have been successfully used in VFM predictions by many authors (see e.g.~\cite{Al-Qutami:2018} and the references therein). However, the forecasting capability of feedforward ANNs is limited because they are unaware of the temporal structure or order between observations. 

Recent results in such applications as automatic text translation and image captioning suggest that the
Long Short-Term Memory (LSTM) model of~\cite{LSTM:1997} is a efficient tool for time series forecasting. 

In order to assess the LSTM model performance for VFM applications, we consider 
a synthetic two-phase severe slugging case (see~\cite{Andrianov:2007}) and  a realistic three-phase well testing dataset. 

For the severe slugging data set, we demonstrate superior performance of LSTM as compared to the feedfoward ANN sliding window approach. We investigate the LSTM convergence as a function of provided distributed pressure measurements and determine the optimal model configuration.

For  the variable rate well test data set, we show that LSTM can successfully handle a noizy dataset, describing a system with changing flow patterns. The accuracy of the forecast improves with the the number of flow periods used for training the model.

\section{LSTM Model Setup}

%$\x(t_0), \x(t_1),\dots,\x(t_E)$
Consider a time series  $\{ \x(t_i) \}$ and $\{ \y(t_i) \}$, where $\x(t_i)$ is a $m$-dimensional  vector of \emph{input features} and $\y(t_i)$ is a $n$-dimensional  vector of \emph{output features}  values at equally spaced time instants $t_i$. In VFM applications, features are the measurement data acquired at different points of the flowline. One can select the sets of the input and output features independently from each other. In particular, a feature can  simultaneously be used for both input and output  (e.g., we might be willing to forecast future values of a flow rate from its past values).

We are interested in forecasting the sequences of output features  of length $l_o$ 
using the sequences of input features  of length $l_i$.
To this end, the  terms $\x(t_i)$ and $\y(t_i)$ from the  \emph{training interval} $[t_0, t_L]$ are divided into $N$ overlapping sequences of length $l=l_i+l_o$, shifted by an indentation step $s$. The result can be cast in form of the training array
\begin{equation}
\label{training_seq}
\X = 
\left[
\begin{array}{ccc}
	\x(t_0)&\dots & \x(t_{l_i-1})  \\
	\x(t_s)&\dots & \x(t_{s+l_i-1})  \\
	\hdotsfor{3} \\
	\x(t_{L-l+1})& \dots & \x(t_{L-l_o})
\end{array}\right]
\end{equation}
and the target array
\begin{equation}
\label{target_seq}
\Y = 
\left[
\begin{array}{ccc}
\y(t_{l_i})  & \dots & \y(t_{l-1})  \\
\y(t_{s+l_i})   & \dots & \y(t_{s+l-1})  \\
\hdotsfor{3} \\
\y(t_{L-l_o+1})   & \dots & \y(t_{L})
\end{array}\right],
\end{equation}
so that $\X\in\Re^{N\times l_i \times m}$ and $\Y\in\Re^{N\times l_o \times n}$.

LSTM maps an input sequence $\x(t_k),\dots, \x(t_{k+l_i-1})$ to the output sequence  $\hat{\y}(t_{k+l_i}),\dots,\hat{\y}(t_{k+l-1})$ for $k=0,\dots,N$ via a composition of linear transformations and nonlinear activation functions. The weights of the linear transformations are 
iteratively updated to minimize a loss function, which penalizes the distance between the output and the target sequences. The original LSTM by~\cite{LSTM:1997} is limited to the case when $l_i=l_o$;~\cite{Cho:2014} and~\cite{Sutskever:2014} introduced an encoder-decoder architecture to generalize the LSTM applicability for cases with $l_i\ne l_o$. See~\cite{Lipton:2015} for a review. 

In this work, we use Keras implementation of LSTM, see~\cite{Chollet:2015}. 
The simulation scripts with the corresponding datasets (see below) are publicly available under \url{https://github.com/nikolai-andrianov/VFM/}.

\section{Experiments}

\subsection{Severe Slugging Case}

Consider a  two-phase isothermal gas-liquid flow in a 60 m section of an offshore pipeline, ending with a 14 m long riser. The flow can be described by a set of partial differential equations, expressing conservation of mass and momentum for the phases. We will be using the mathematical model, numerical method, and the specifications for the test case, presented in~\cite{Andrianov:2007}.

\begin{figure}[h]
	\begin{center}
		\includegraphics[width=0.8\textwidth]{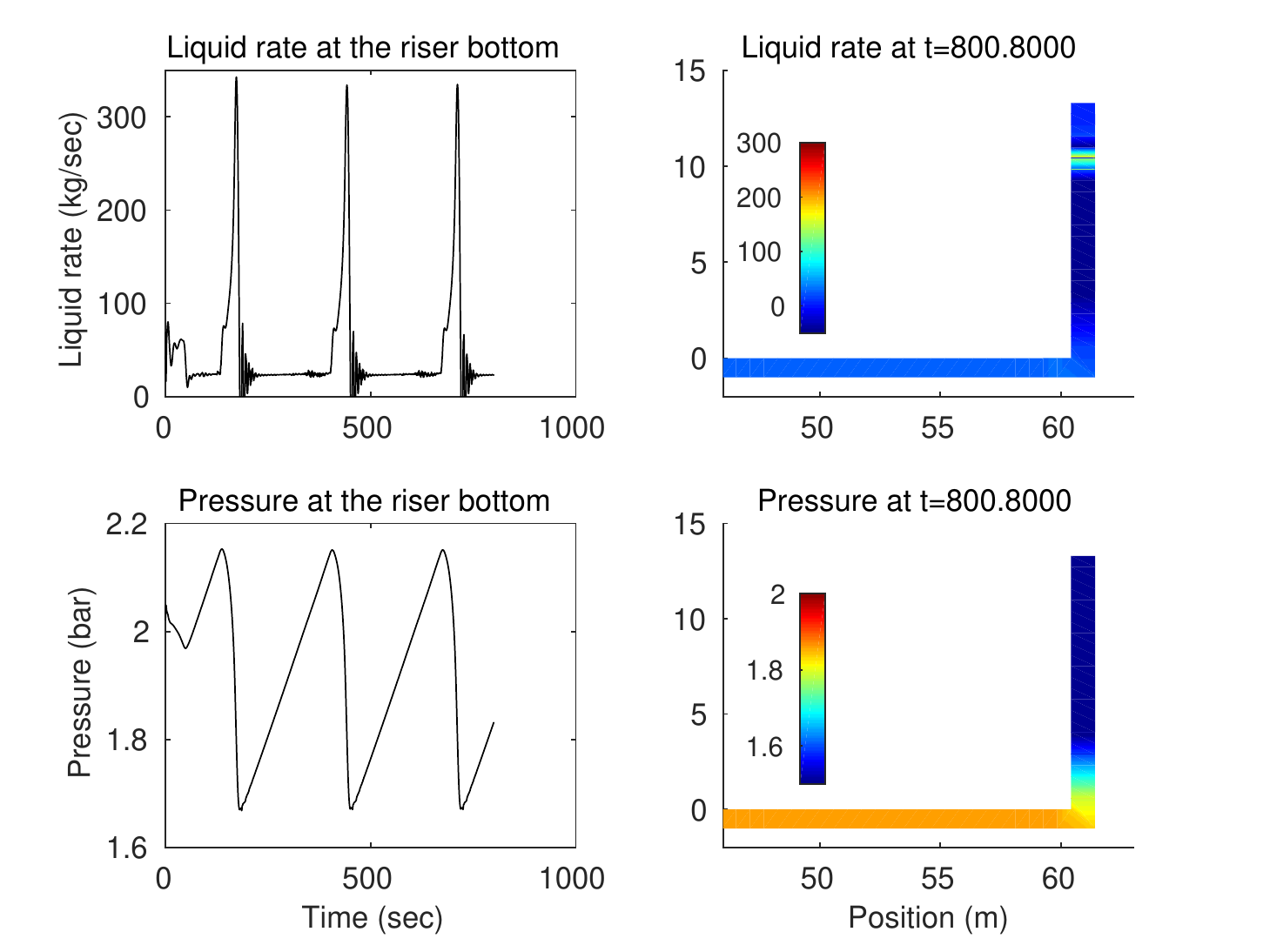}    
		\caption{A snapshot of the numerical solution for the severe slugging case at intermediate time instant.} 
		\label{fig:riser}
	\end{center}
\end{figure}

Under certain constant boundary conditions at the pipeline inlet and at the riser outlet, the numerical solution exhibits a typical severe slugging behaviour, see Fig.~\ref{fig:riser}.

\begin{figure}[t]
	\begin{center}
		\includegraphics[width=0.8\textwidth]{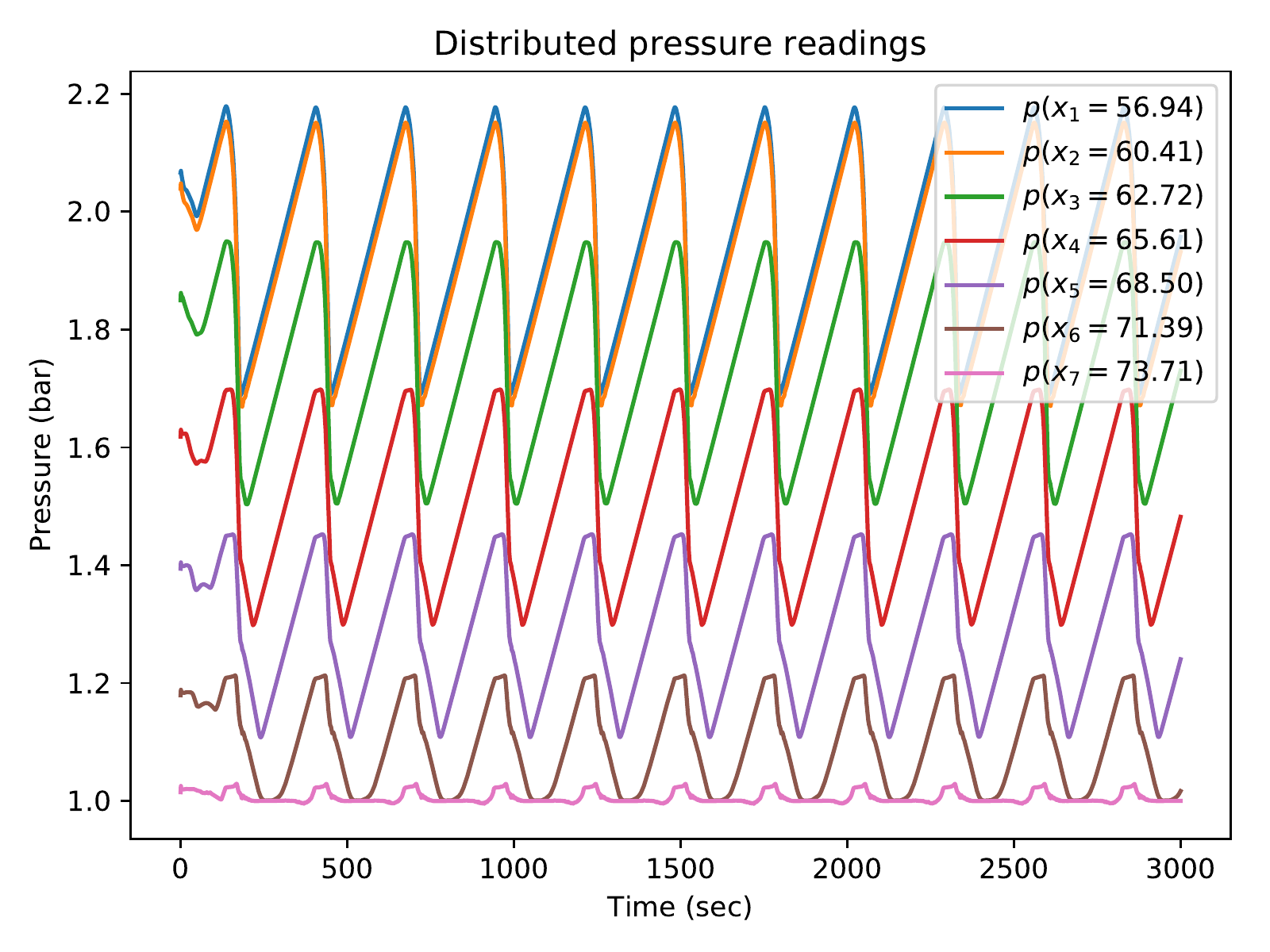}    
		\caption{Pressure data used as input to forecast the flow rates.} 
		\label{fig:riser_pressure}
	\end{center}
\end{figure}

We will utilize this numerical solution as a ``ground truth'' for  forecasting the liquid and gas rates at the riser bottom using the data from  virtual pressure gauges distributed along the flowline, see Fig.~\ref{fig:riser_pressure}.

In order to run LSTM forecasts, we   resample the normalized pressure and flow rate data with a uniform timestep of 1 sec, and use half of the total hydrodynamical simulation time  as a training interval, $[t_0, t_L]=[0, 1500]$ sec. 

We first analyze the quality of LSTM forecasts when the network is trained using only pressure readings as input and liquid rate as output. The training data is divided into $N=1127$ sequences of length $l=374$ sec with $l_i=l_o$, shifted by the indentation step $s=1$ sec. The network details are given below:
\begin{itemize}
	\item Deep LSTM with 3 hidden layers and 10 memory cells at each layer;
	\item Total number of trainable parameters is $2171,\dots, 2411$ with validation split of 0.05 for the number  of input features $m=1,\dots,7$ and number  of output features $n=1$;
	\item Fixed random seed for repeatability in parameter initialization;
	\item Mean squared error (MSE) loss function and Adam optimizer of~\cite{Kingma:2014} with batch size of 1 and number epochs equal to 10.
\end{itemize}
These network training parameters were determined by trial-and-error. For the case considered, the forecasting results were most sensitive to the number and lengths of input/output sequences.

\begin{figure}[h]
	\begin{center}
		\includegraphics[width=0.8\textwidth]{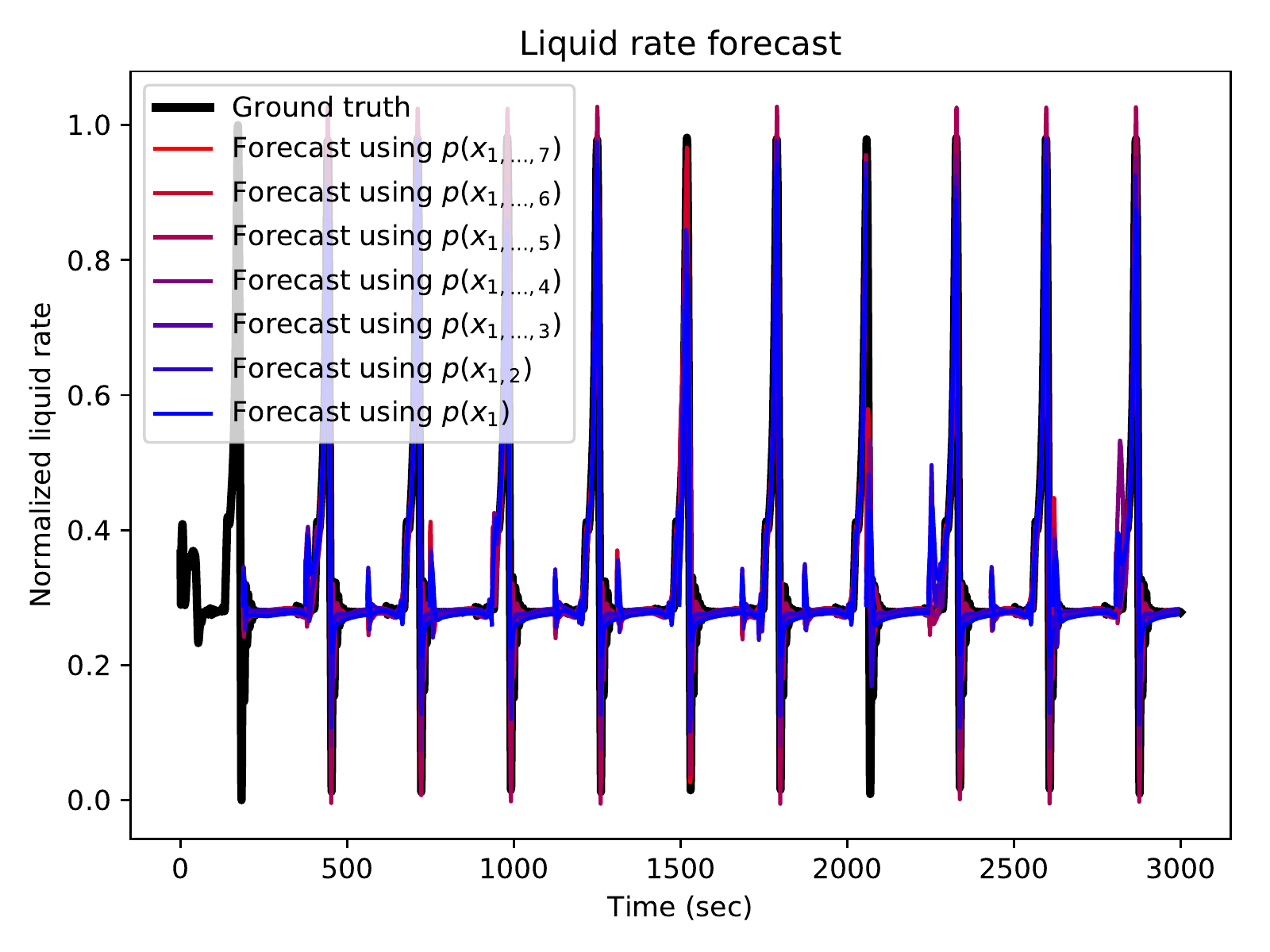}    
		\caption{LSTM liquid rate forecasts using various number of pressure readings as input.} 
		\label{fig:pres_riser_pq}
	\end{center}
\end{figure}

\begin{figure}[H]
	\begin{center}
		\includegraphics[width=0.8\textwidth]{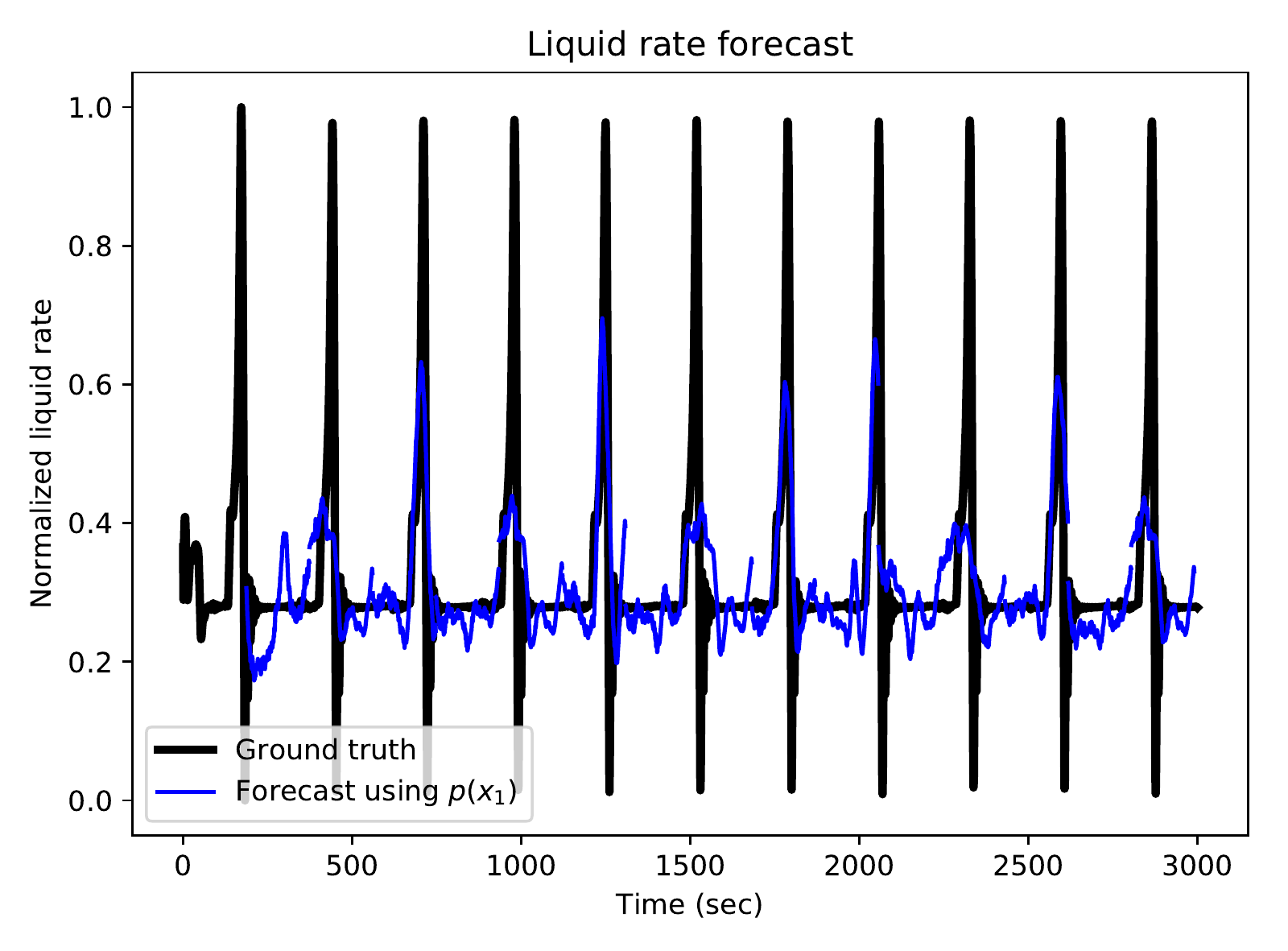}    
		\caption{Feedforward ANN liquid rate forecasts using a single  pressure readings as input.} 
		\label{fig:FF_pres_riser_pq}
	\end{center}
\end{figure}

The forecasting capability of an LSTM can be quantified with the ratio
\begin{equation}
f = \frac{l_o}{t_L-t_0}\cdot \mbox{100}\%,
\end{equation}
which we will term the \emph{relative forecasting interval}. For the severe slugging case $f=12.4\%$, i.e.\ the  LSTM can forecast the future flow rates 
for the time interval which length is  $12.4\%$ of the length of LSTM's training interval.

The forecasts are plotted as 15 \emph{non-overlapping} sequences of length 
$l=374$ sec with $l_i=l_o=187$ sec, shifted by the indentation step $s=187$ sec, see Fig.~\ref{fig:pres_riser_pq}. Observe that even when trained on a single pressure reading, LSTM yields excellent agreement with the ground truth hydrodynamical solution in terms of the frequency and amplitude of the liquid rate peaks. This is in striking contrast to the results of a feedforward  ANN using sliding window  approach with  3 hidden layers and 10 neurons at each layer, trained on the same dataset as the LSTM, see Fig.~\ref{fig:FF_pres_riser_pq}.

\begin{figure}[t]
	\begin{center}
		\includegraphics[width=0.8\textwidth]{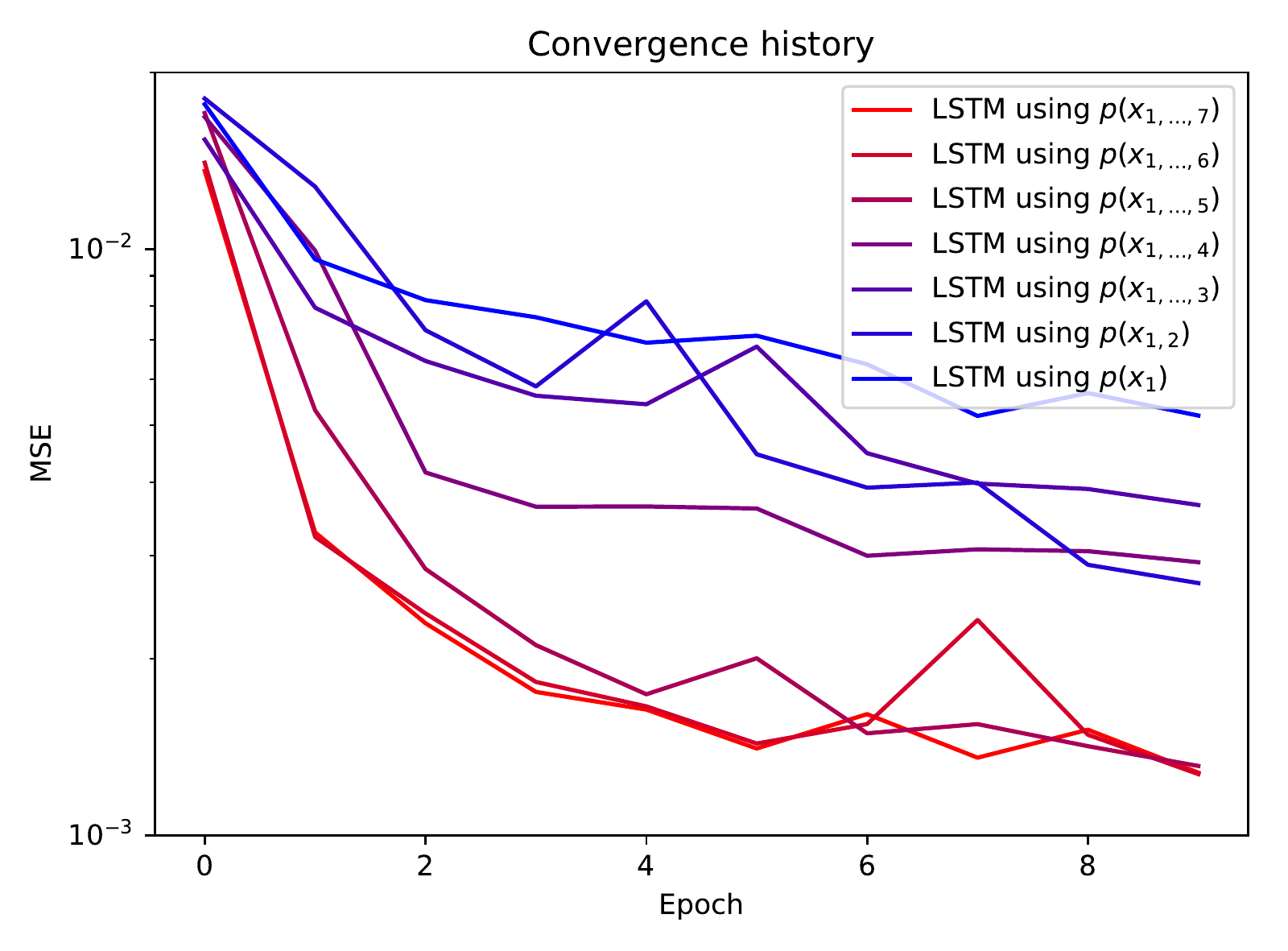}    
		\caption{LSTM convergence history as a function of number of pressure readings used to train  the network.} 
		\label{fig:pres_riser_conv}
	\end{center}
\end{figure}

Adding more pressure data as the training input does generally increase the accuracy of LSTM forecasts. However, this improvement is not monotonous, and starting from a certain number of pressure readings (in this case 5 readings) the accuracy remains essentially the same, see Fig.~\ref{fig:pres_riser_conv}.

\begin{figure}[t]
	\begin{center}
		\includegraphics[width=0.8\textwidth]{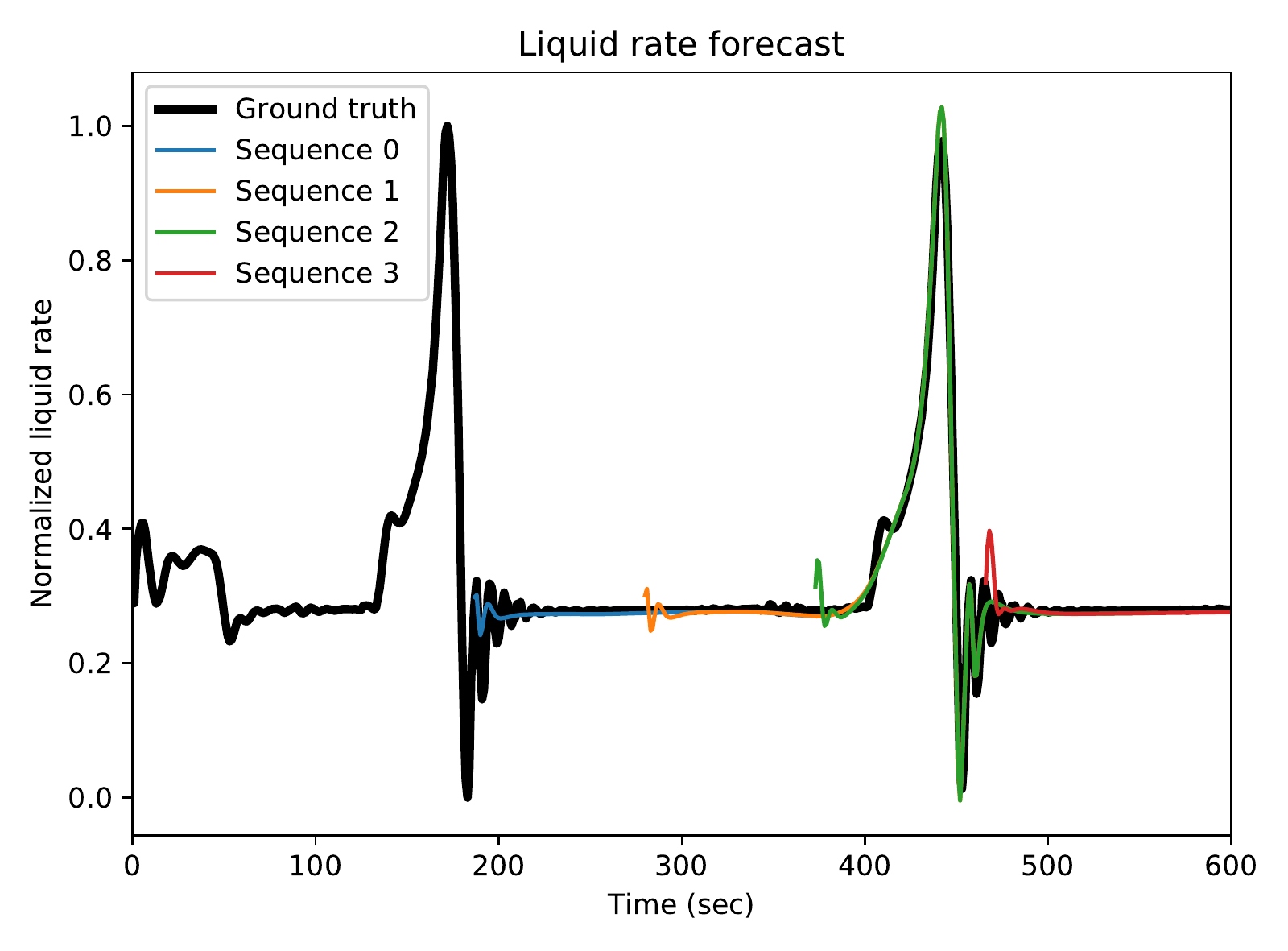}   
		\caption{First output sequences of zoomed LSTM liquid rate forecasts using 5 pressure readings as input.} 
		\label{fig:pres_riser_pq_zoom}
	\end{center}
\end{figure}

Note that there are spurious oscillations visible in LSTM forecasts on  Fig.~\ref{fig:pres_riser_pq}. 
We replot the zoomed LSTM forecasts using 5 pressure readings as \emph{overlapping} sequences of the same length $l_i=l_o=187$ sec, but shifted by the indentation step $s=93$ sec, see Fig.~\ref{fig:pres_riser_pq_zoom}. 

Observe that the spurious oscillations are located at the beginning of each sequence. This is not surprising because LSTM learns its weights within the input sequence. However, these oscillations do not affect the accuracy of the forecasts if we use overlapping output sequences. Indeed, referring to Fig.~\ref{fig:pres_riser_pq_zoom} we have by the end of $t=l_i$ the 0\ts{th} sequence forecast till $t=2l_i$, which is oscillation-free by the time $t=l_i+s$, when the new 1\ts{st} sequence forecast is made till  $t=2l_i+s$. We keep using the 
0\ts{th} sequence forecast until the oscillations in the 1\ts{st} sequence forecast disappear, and repeat the process.

\begin{figure}[t]
	\begin{center}
		\includegraphics[width=0.8\textwidth]{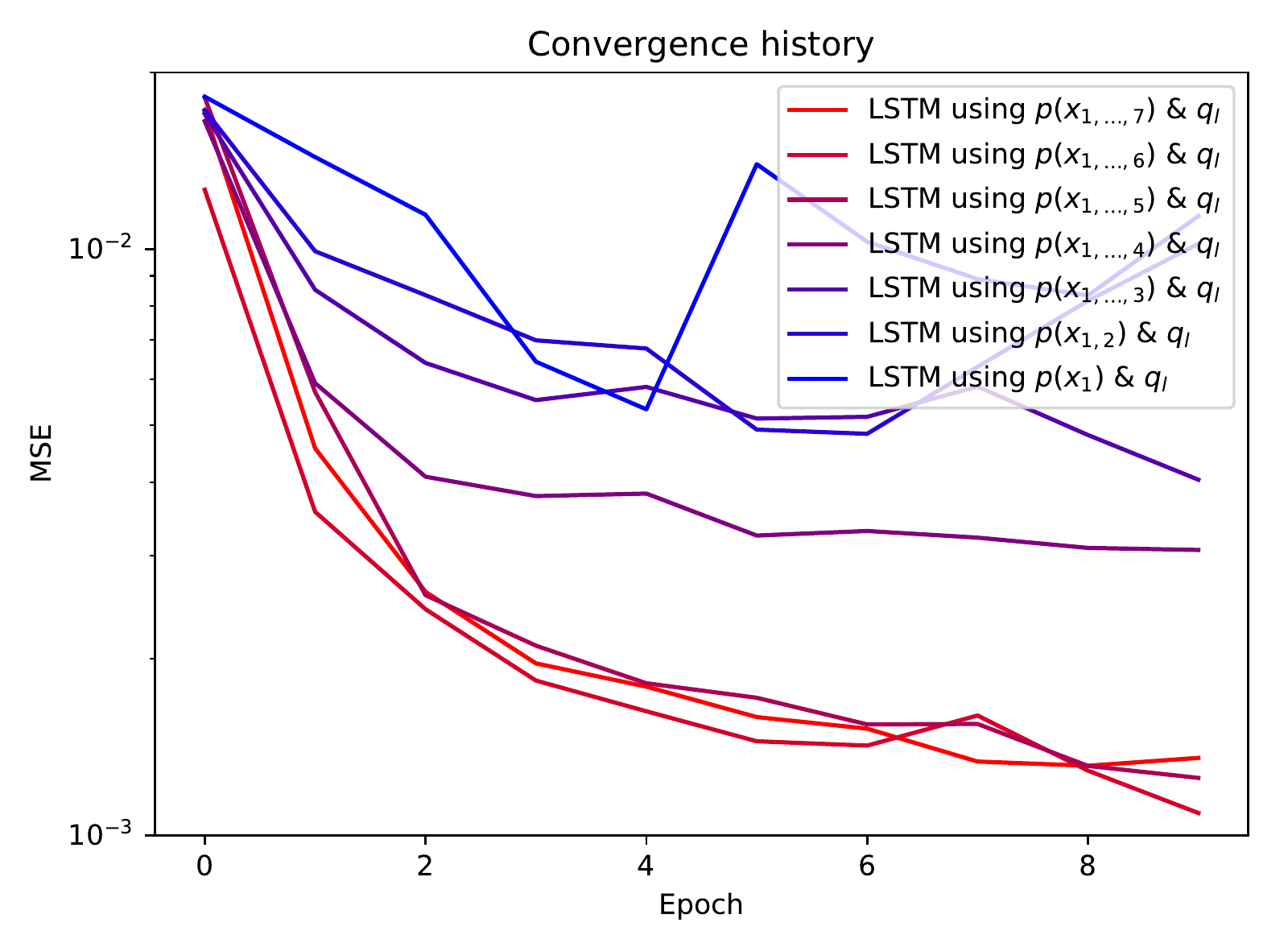}    
		\caption{Convergence history of LSTM trained on several pressure readings and the liquid rate.} 
		\label{fig:pres_ql_riser_conv}
	\end{center}
\end{figure}

The performance of LSTM  trained on pressure \emph{and} liquid rate is presented in Fig.~\ref{fig:pres_ql_riser_conv}. Observe that increasing the  number of  measurements used to train the network does not improve the accuracy of the forecast, cf.\ Fig.~\ref{fig:pres_riser_conv}. Moreover, if few pressure readings are used for training, the performance of LSTM  trained on pressure and rate data becomes worse than that of LSTM trained just on pressure data.

The accuracy of LSTM forecasts of both liquid and gas rates (i.e., $n=2$ output features) is essentially the same as the results presented above for liquid rate forecasts only.

We also tested the encoder-decoder LSTM of~\cite{Cho:2014} and~\cite{Sutskever:2014}, but the forecasts were less accurate  compared to the results presented above.

Wall time required for  training of LSTMs described above with was approx. 30 mins using a single core of i7-7700HQ CPU. Using 8 cores of the same CPU resulted in approx.\ 20\% speedup.

\subsection{Variable Rate Well Test}

\begin{figure}[h]
	\begin{center}
		\includegraphics[width=0.8\textwidth]{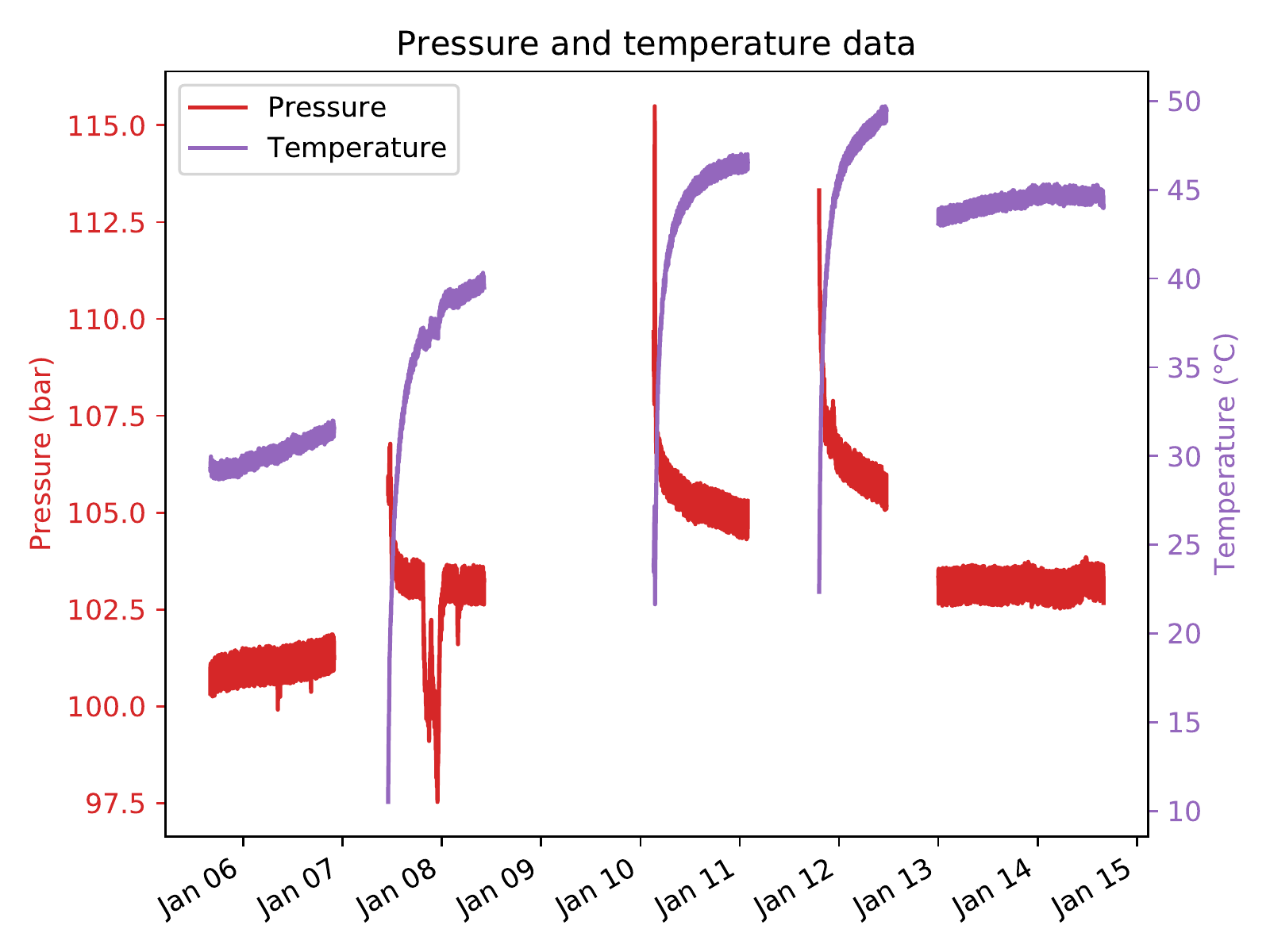}    
		\caption{Pressure and temperature for the variable rate well test.} 
		\label{fig:wt_PT}
	\end{center}
\end{figure}

\begin{figure}[H]
	\begin{center}
		\includegraphics[width=0.8\textwidth]{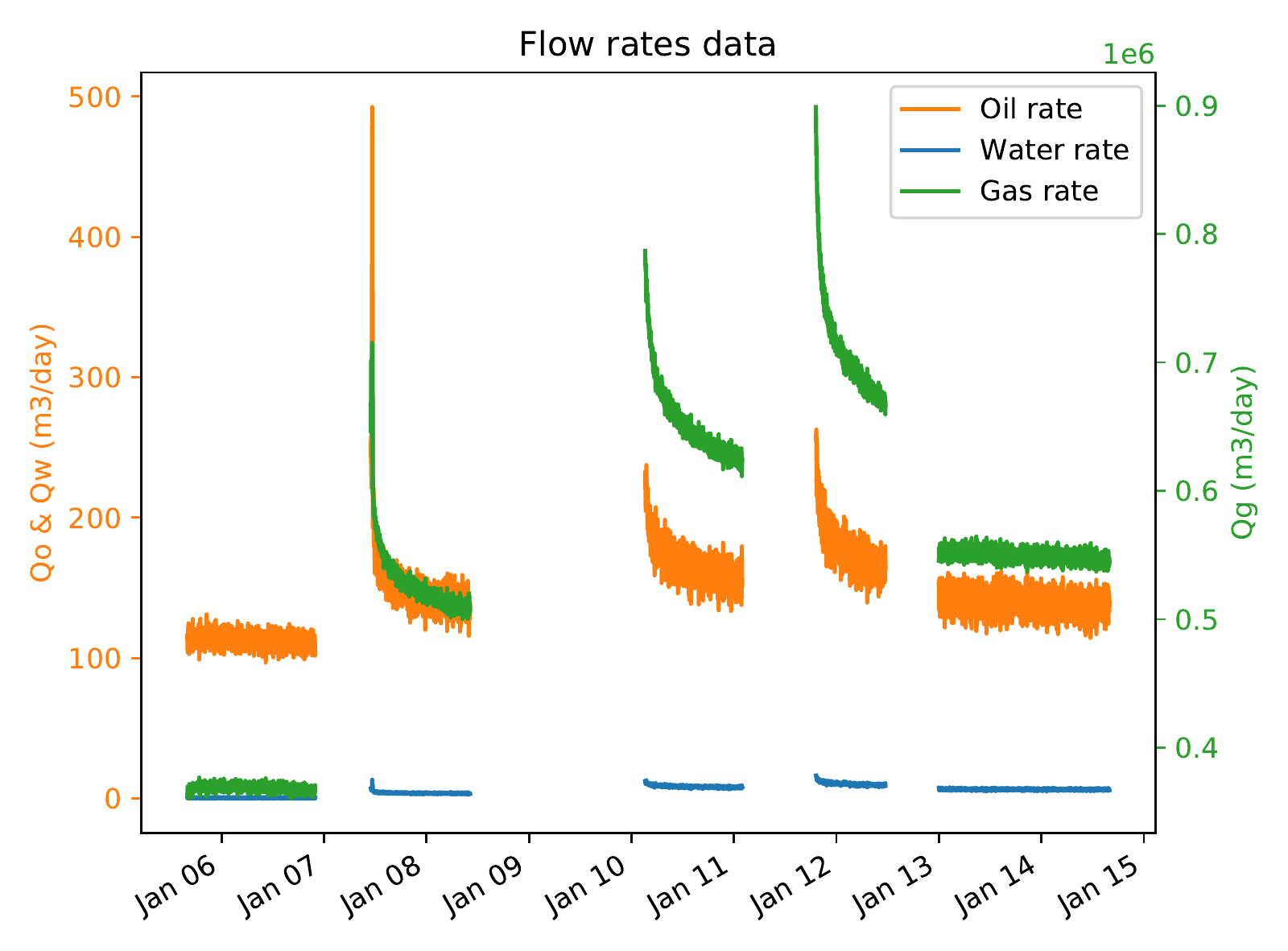}    
		\caption{Multiphase rates for the variable rate well test.} 
		\label{fig:wt_Q}
	\end{center}
\end{figure}

Consider a synthetic dataset of pressure, temperature, and oil, gas and water rates 
measurements during a well test, see Fig.~\ref{fig:wt_PT} and  Fig.~\ref{fig:wt_Q}.
The data is characteristic for a rich gas condensate \emph{deliverability} test, which involves flowing the well on successively larger choke sizes in order to determine the well's inflow performance relationship (IPR) and maximize gas condensate recovery. (In what follows, we will refer to gas condensate  as ``oil''.)

The dataset consists of 5 flow periods, which are characterized by the corresponding choke size. Within each flow period, the measurements are generally sampled with the uniform timestep of 1 min.
We are interested in forecasting the multiphase rates using the values of pressure and temperature.

\begin{figure}[t]
	\begin{center}
		\includegraphics[width=0.8\textwidth]{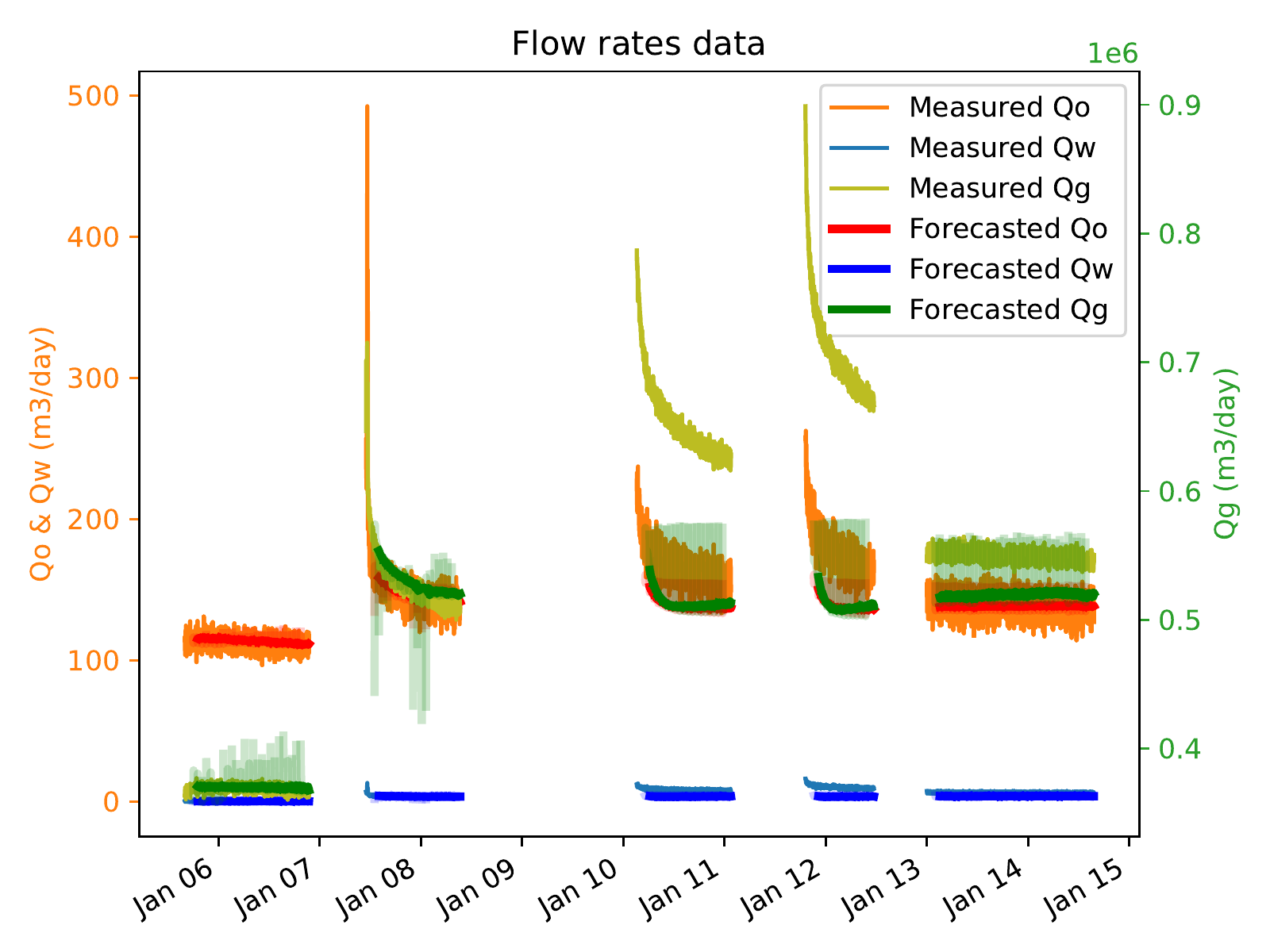}    
		\caption{Multiphase rates forecast for the LSTM trained on first 2  flow periods. Spurious oscillations are plotted semi-transparent. } 
		\label{fig:wt_Q_forecast_FP01}
	\end{center}
\end{figure}

To this end, we will be utilizing essentially the same procedure as for the severe slugging experiment. We train the network on first flow periods
using pressure and temperature readings as input features and multiphase rates as output feactures. Then, the multiphase rates forecasts are run for all flow periods.

One key difference of the variable rate well test case from the severe slugging case considered in the previous section is that the flow pattern in the well test case changes drastically from one flow period to another, see Fig.~\ref{fig:wt_PT} and  Fig.~\ref{fig:wt_Q}. This constitutes a challenge to the neural network, because we try to approximate the behaviour of the \emph{changing} flow system with the same ANN. Another difference and a challenge for the neural network is that the dataset is noizy.

In what follows, we will compare the forecasting accuracy of LSTMs trained on first 2 and 3 flow periods. For these two cases, the training data  is divided into $N=2705$
and $N=3828$ sequences, respectively. In both cases the sequence length is 
 $l=244$ min with $l_i=l_o$, and the indentation step is $s=1$ min. 
The relative forecasting intervals are $f=3.8\%$ and $f=2.6\%$, respectively. 
 The LSTM structure is the same as described in the previous section. The forecasts are presented in sequences of  $l=244$ min with the indentation step is $s=l_i/2$ min.

The results for the LSTM, trained on  first 2  flow periods, are presented in  Fig.~\ref{fig:wt_Q_forecast_FP01}.

\begin{figure}[t]
	\begin{center}
		\includegraphics[width=0.8\textwidth]{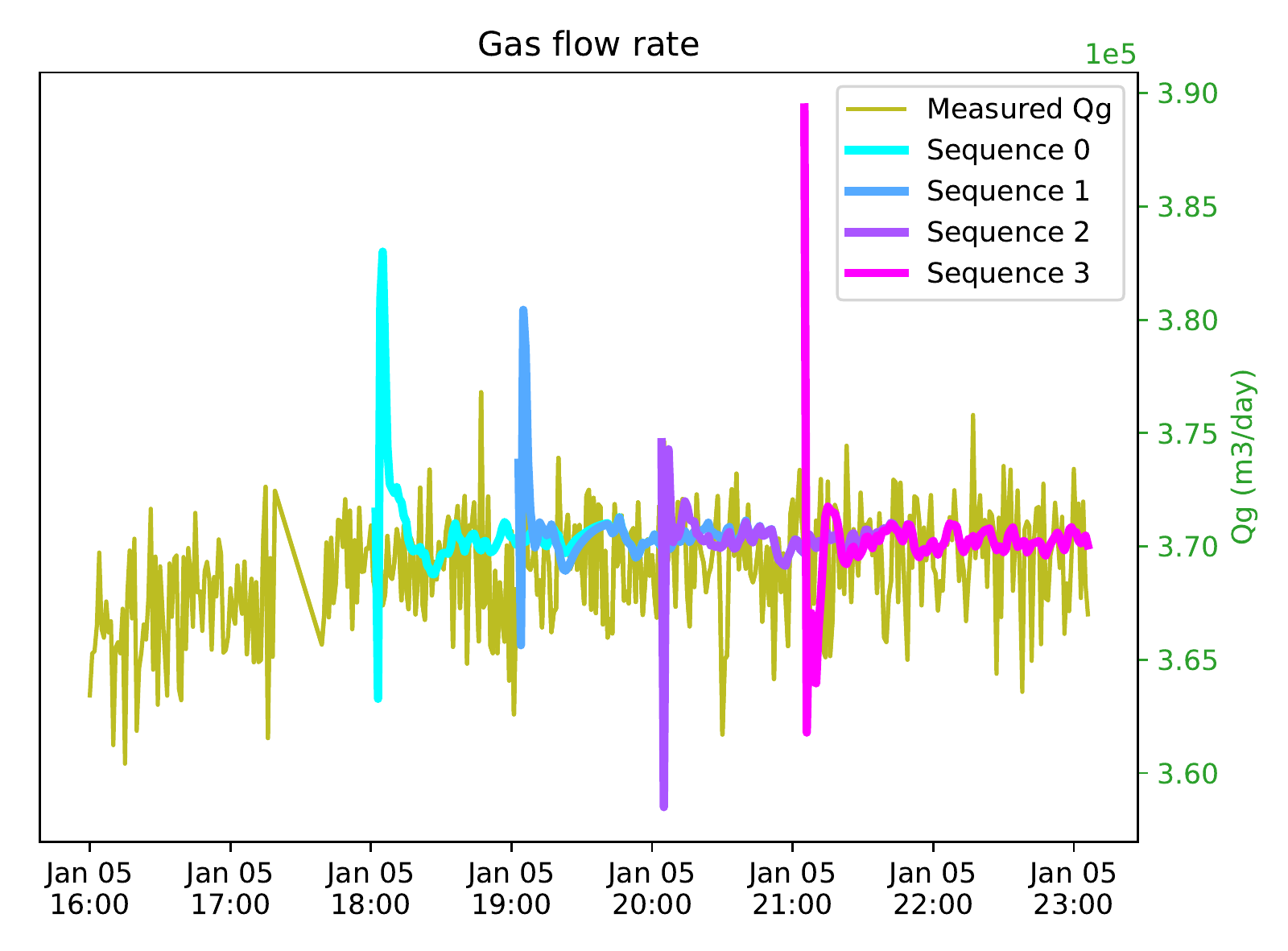}   
		\caption{First output sequences of zoomed LSTM gas rate forecasts.} 
		\label{fig:wt_Qg_forecast_zoomed}
	\end{center}
\end{figure}

The model reproduces well the training data from the first 2 flow periods. 
The best accuracy is achieved for forecasted values of oil and water rates, while the gas rate is slightly overestimated. Still, the trends for all rates are captured correctly. 

On the testing set (flow periods 3 to 5), the model yields reasonable values for the oil and water rates. However, the forecasts for gas rate are non-satisfactory. This can be explained by the fact that both oil and water rates lie in a same range throughout all flow periods, which is not the case for the gas rate. Also, note that the first $l_i$ data points on each training sequence are not covered by any output sequence. Consequently, the sharp peaks at the beginning of the flow periods 2, 3, and 4 are not included in the training dataset.

On Fig.~\ref{fig:wt_Q_forecast_FP01} we witness the same spurious oscillations as discussed in the previous section, cf.\ Fig.~\ref{fig:pres_riser_pq_zoom}. To see this, in Fig.~\ref{fig:wt_Qg_forecast_zoomed} we plot the measured gas rate together with the first output sequences of forecasted gas rate during the 1\ts{st} flow period. Observe that the peaks are located at the beginning of each output sequence of length $l_o=122$ min. By following the same reasoning as in the previous section, we argue that these  spurious oscillations do not affect the quality of the forecast.

\begin{figure}[t]
	\begin{center}
		\includegraphics[width=0.8\textwidth]{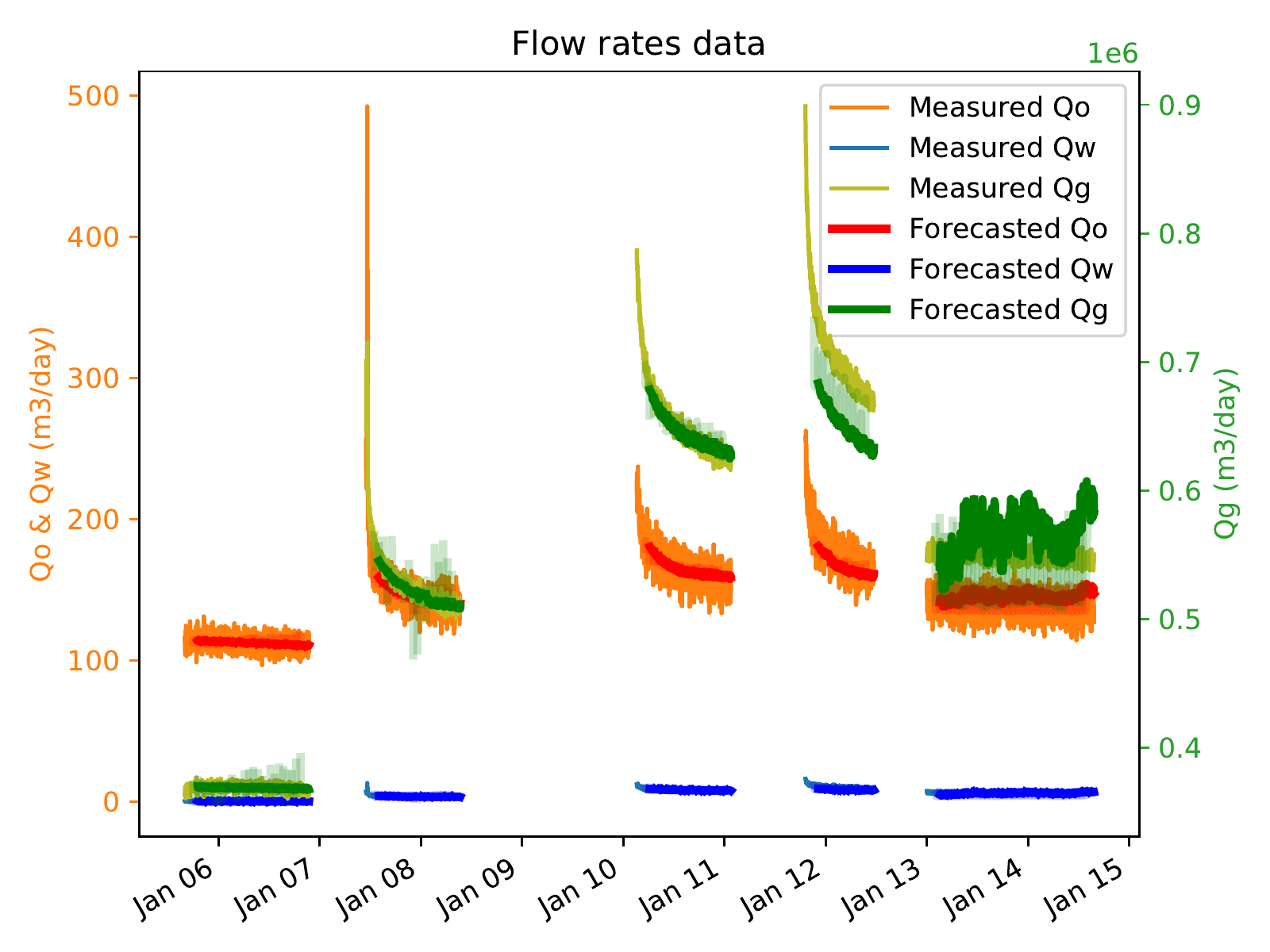}    
		\caption{Multiphase rates forecast for the LSTM trained on first 3  flow periods. Spurious oscillations are plotted semi-transparent. } 
		\label{fig:wt_Q_forecast_FP012}
	\end{center}
\end{figure}

The results for the LSTM, trained on  first 3  flow periods, are presented in  Fig.~\ref{fig:wt_Q_forecast_FP012}.
The model reproduces well all training data from the first 3 flow periods. 
It is interesting to note that the forecasts are less noizy as compared to the measured data. There are spurious oscillations visible on the graphs, but their amplitude is less than that of the model, trained on 2 flow periods.

On the testing set (flow periods 4 and 5), the model performance is best for oil and water rates, and less satisfactory for gas rates. Again, this can be explained by a larger variability of the gas rate as compared to oil and water rates. Overall, the accuracy of the forecast is considerably better compared to that of the model, trained just on 2 flow periods.

Wall time required for  training of LSTMs on first  two and three flow periods  using a single core of i7-7700HQ CPU  was approx. 50  and 90 min, respectively.

\section{Conclusion}

In this work, we have shown that LSTM can be considered as a robust tool for forecasting the values of multiphase rates using pressure and temperature data. The best accuracy was achieved when the lengths of the input and output sequences to LSTM were equal. Consequently, we are limited in the length of the time interval suited for forecasts. Removing this limitation without sacrifice on the accuracy of the forecast can be an interesting topic of future research.

%\begin{ack}
%Place acknowledgments here.
%\end{ack}

\bibliography{ifacconf}             % bib file to produce the bibliography
                                                     % with bibtex (preferred)
                                                   
\bibliographystyle{siam}

\end{document}